\documentclass[conference]{IEEEtran}

\usepackage{cite}
\usepackage{amsmath,amssymb,amsfonts}
\usepackage{algorithmic}
\usepackage{graphicx}
\usepackage{textcomp}
\usepackage{xcolor}
\usepackage{multirow}
\usepackage[hyphens]{url}
\usepackage{verbatim}
\usepackage{tabularx}
\usepackage{booktabs}
\usepackage{comment}
\usepackage{tikz}
\usepackage{enumitem}
\usepackage{pgfplots}
\pgfplotsset{compat=1.18}

\title{Multi-Task Learning for Heterogeneous Prediction from Video Game State with Transfer Learning}

\iftrue

\author{\IEEEauthorblockN{Jonas Pech\'e}
    \IEEEauthorblockA{\textit{Computer Graphics} \\
    \textit{Johannes Kepler University Linz}\\
    Linz, Austria \\
    j\_peche@wargaming.net}
\and
    \IEEEauthorblockN{Aliaksei Tsishurou}
    \IEEEauthorblockA{\textit{Independent Researcher} \\
    Wroclaw, Poland \\
    aliakseitsishurou@gmail.com}
\and
    \IEEEauthorblockN{Alexander Zap}
    \IEEEauthorblockA{\textit{DS Research} \\
    \textit{Wargaming}\\
    Berlin, Germany \\
    a\_zap@wargaming.net}
\and
    \IEEEauthorblockN{G\"unter Wallner}
    \IEEEauthorblockA{\textit{Computer Graphics} \\
    \textit{Johannes Kepler University Linz}\\
    Linz, Austria \\
    guenter.wallner@jku.at}
}

\else

\author{\IEEEauthorblockN{Anonymous Authors}}

\fi

\IEEEoverridecommandlockouts

\newif\ifarxiv
\arxivtrue       

\newcommand{\IEEEAcceptedManuscriptNotice}{%
  \textcopyright{} 2026 IEEE.
  Personal use of this material is permitted.
  Permission from IEEE must be obtained for all other uses,
  in any current or future media, including reprinting/republishing
  this material for advertising or promotional purposes,
  creating new collective works, for resale or redistribution
  to servers or lists, or reuse of any copyrighted component
  of this work in other works.%
}

\begin{document}

\ifarxiv
\else
  \IEEEpubid{%
    \makebox[\columnwidth]{%
      979-8-3315-9476-3/26/\$31.00~\copyright2026 IEEE
      \hfill
    }%
    \hspace{\columnsep}\makebox[\columnwidth]{}%
  }
\fi

\maketitle

\ifarxiv
  \begingroup
    \renewcommand{\thefootnote}{}
    \footnotetext{\scriptsize\IEEEAcceptedManuscriptNotice}
  \endgroup
\else
  \IEEEpubidadjcol
\fi

\begin{abstract}
Multi-task learning (MTL) is a promising approach for prediction tasks derived from video game state data, as modern game telemetry provides multiple related supervision signals from the same structured observations. We study whether a shared model trained jointly across tasks in team-based multiplayer games can improve generalization while reducing training and inference cost compared to specialized single-task models. We adapt a multimodal architecture for endpoint prediction to a general multi-task setting that combines rasterized vision inputs, global match context, and per-unit state information through an image encoder and attention-based interaction modeling. Experiments on a large proprietary \emph{World of Tanks} dataset compare single-task and multi-task training, evaluate weighting strategies for mixed losses and conflicting gradients, and test pre-training/fine-tuning under limited target-data regimes. We also examine within-game transfer across game maps under structured environment shift. 
\end{abstract}

\section{Introduction}

Multi-task learning (\emph{MTL}) is an effective approach for learning related prediction tasks jointly with a single model~\cite{ruder2017overviewmultitasklearningdeep}. Learning related tasks jointly encourages a shared representation that can generalize better than training each task in isolation, while also improving model size, training efficiency, and inference efficiency by replacing several models with one~\cite{caruana1997multitask,ruder2017overviewmultitasklearningdeep}. In supervised settings, MTL relies on the availability of multiple labels for the same observations, allowing a shared encoder to be optimized from several related objectives at once. A common practical design is \emph{hard parameter sharing}, in which a single shared encoder is trained jointly across tasks and only task-specific output heads remain separate~\cite{caruana1997multitask,ruder2017overviewmultitasklearningdeep}.

Video games are a promising application domain for this setup because modern telemetry pipelines provide rich structured state observations together with multiple labels derived from the same gameplay data. In practice, predictions from such data can support downstream game AI and analytics components, for example, for positioning support, threat estimation, or base-capture prevention. Joint optimization also aligns well with production requirements such as reducing compute costs during both training and inference, while shared representations can be beneficial when learning related heterogeneous targets~\cite{Vandenhende_2021}.

Particularly in team-based games, the same prediction tasks are often observed under varying environments such as maps that preserve the rules and labels but change the input distribution through terrain, visibility, and movement constraints. Cross-map transfer is therefore a natural extension of the MTL setting. While this does not test transfer across different games or rule systems, it is of practical relevance when data on new or unreleased maps is limited. 

In this paper, we study supervised multi-task prediction from video game states in a hard parameter sharing setup, including transfer across different game maps. Building on prior work on multimodal endpoint prediction for \emph{World of Tanks}~\cite{peche2025multimodalarchitectureendpointposition}, we adapt the architecture to a broader multi-task setting over heterogeneous game state targets. Our contribution is an applied empirical study of existing MTL and transfer-learning techniques, rather than a new MTL optimizer or architecture. We consider the following research questions: 

\begin{description}
    \item[\textbf{RQ1:}] Does MTL improve performance on representative supervised prediction tasks derived from video game state compared to single-task learning?
    \item[\textbf{RQ2:}] How well do representative MTL weighting and gradient-balancing algorithms perform on heterogeneous tasks with mixed output modalities and loss functions?
    \item[\textbf{RQ3:}] Can source-task pre-training improve target-task fine-tuning across different target-data regimes?
    \item[\textbf{RQ4:}] Does the shared representation learned by MTL remain useful across different game maps of the same game?
\end{description}

\noindent To answer these questions, we first describe the dataset and define the corresponding prediction tasks in Section~\ref{sec:data}. Then we adapt a generic multimodal model framework suited to team-based games and MTL settings~\cite{peche2025multimodalarchitectureendpointposition}, described in Section~\ref{sec:methodology}. Finally, we analyze the effect of different MTL weighting strategies, evaluate source-task pre-training for target-task adaptation under varying data availability, and test within-game cross-map transfer in Section~\ref{sec:evaluation}. 

\section{Related Work}
    \label{sec:related_work}

MTL has long been studied as a framework for improving generalization through shared representations across related tasks~\cite{caruana1997multitask}. In the following, we briefly review prior work on multi-task optimization, multimodal game-state architectures, and learning from game environments.

\subsection{Balancing Objectives in Multi-task Optimization}
The main risk of MTL is negative transfer, where jointly trained tasks can degrade performance when they have conflicting gradients in the shared parameters~\cite{yu2020gradient}. Empirical studies, mostly in vision but also in general applications, point out that success depends heavily on task relatedness, the amount of pattern and information overlap between tasks, and the optimization strategy used for learning~\cite{Abdelsamie2025}. This paper at hand focuses on mixed losses (classification and regression) with a shared feature extractor. Such a setting is especially susceptible to this problem and requires more careful design of the optimization and balancing strategies~\cite{kendall2018multitasklearningusinguncertainty}.

The fundamental approach to balance different loss functions is by weighting per-task losses with individual scalar weights. Uncertainty weighting, for example, uses probabilistic estimates of task uncertainty to scale each loss, giving less weight with higher uncertainty~\cite{kendall2018multitasklearningusinguncertainty}. More recently, analytical experiments and refinements compute task weights directly from per-task losses using a normalized softmax-based function, that is both more stable and computationally cheaper on the evaluation datasets~\cite{kirchdorfer2024analyticaluncertaintybasedlossweighting}.

Slightly more advanced approaches consider the gradients of individual tasks as well. Examples are: \emph{GradNorm} balances training by normalizing the gradient magnitudes across tasks by dynamically adjusting task-weights to match the target learning rate~\cite{chen2018gradnormgradientnormalizationadaptive}. MTL can also be seen as a multi-objective optimization problem: the \emph{Multiple Gradient Descent Algorithm} (MGDA) finds the convex combination of all task gradients that represents a shared descent direction, reducing all task losses at once~\cite{desideri:inria-00389811}. \emph{PCGrad} proposed ``gradient surgery'', projecting away conflicting directions when the task gradients have negative dot products~\cite{yu2020gradient}. \emph{CAGrad} further regularizes optimization to avoid local task degradation and reported further improvements in challenging multi-task environments~\cite{liu2021cagrad}. Fast Adaptive Multitask Optimization (FAMO) targets the $\mathcal{O}(T)$ cost of methods requiring the individual gradients for each task and proposes a dynamic weighting scheme trained using a second optimizer~\cite{liu2023famo}.

It is also worthwhile to mention random loss weighting (RLW), which trivially samples weights at random while often matching or even outperforming other, sometimes even more sophisticated, approaches on a subset of datasets~\cite{lin2022reasonableeffectivenessrandomweighting}, highlighting the importance of testing whether the current state-of-the-art method actually works well with the given task and data definitions.

Most weighting methods treat each task independently, but there are also sample-level approaches such as \emph{SLGrad}, which reweights individual training examples based on their contribution. \emph{SLGrad} mainly focuses on auxiliary task setups, where such tasks may have little or even no relevance in a fraction of samples, but the approach itself applies to other settings as well~\cite{gregoire2023samplelevelweightingmultitasklearning}. In this paper, we focus on a representative subset of task-level balancing methods: equal weighting, RLW, FAMO, and \emph{PCGrad}.

\subsection{Multi-task Architectures for Structured and Visual Game-state Representations}

For visual game-state inputs, convolutional encoders are a common choice for extracting spatial context from rasterized observations. \emph{EfficientNet}-style encoders are often used where efficiency matters~\cite{tan2020efficientnetrethinkingmodelscaling}, and transformer-based vision models have recently achieved high embedding quality for feature extraction~\cite{dosovitskiy2021imageworth16x16words}. A study outside of the games domain shows that combining visual representations with classification and regression objectives in an MTL setting can improve performance, provided that the associated optimization challenges are handled appropriately~\cite{Vandenhende_2021}.

Modern game-state representations often contain variable-sized sets of entities and benefit from attention mechanisms to model communication and interactions~\cite{vaswani2023attentionneed}. For instance, \emph{AlphaStar} processes a list of units with self-attention and combines spatial and other signals, illustrating the relevance of attention-based encoders in complex multi-agent environments~\cite{vinyals2019grandmaster}. More generally, combining structured entity features with visual context is a useful design pattern in game-state modeling, enabling models to capture both local spatial structure and global interaction information~\cite{vinyals2019grandmaster,peche2025multimodalarchitectureendpointposition}. These works motivate the input representation and encoder design, while our focus is supervised multi-task prediction with heterogeneous output heads.

\subsection{Learning from Game Environments}

Deep Q-Learning (DQN) showed that deep neural networks can learn a policy from raw pixel observations, introducing deep reinforcement learning into games~\cite{mnih2015human}. Subsequent work in this direction, such as \emph{IMPALA}, solved multiple tasks using a single agent and reported positive knowledge transfer between tasks in video game environments~\cite{espeholt2018impalascalabledistributeddeeprl}. Methods such as \emph{Distral} further improved multi-task reinforcement learning by distilling a jointly trained policy to stabilize training and mitigate negative transfer~\cite{teh2017distralrobustmultitaskreinforcement}. Varghese and Mahmoud~\cite{Varghese} presented a framework in which multiple workers each learn a single task while a shared network enables knowledge transfer across tasks. More recent approaches such as \emph{Bootstrap Your Own Teacher}~\cite{11114408} extend this idea by performing online policy distillation during training, allowing agents to iteratively refine shared knowledge across multiple games without requiring a fixed teacher policy. Sequence-modeling approaches such as \emph{Decision Transformers} have provided an alternative perspective on learning control policies from offline trajectories. However, these methods target return-conditioned sequential decision-making rather than supervised multi-task prediction from a single game-state snapshot and are therefore outside the main scope of this paper~\cite{chen2021decisiontransformerreinforcementlearning}. Our focus is instead on supervised multi-task prediction and transfer from existing datasets.

Publicly available datasets, such as \emph{ESTA}~\cite{xenopoulos2022estaesportstrajectoryaction} for \emph{CS:GO}, enable research and benchmarking on game prediction tasks, and studies on similar games -- such as \emph{Knowledge Enhanced Graph Contrastive Learning} for MOBA match outcome prediction~\cite{JIANG2025104010} -- have explored single-task modeling with auxiliary tasks to improve the main prediction. However, these works typically focus on a single primary objective, rather than studying general-purpose multi-head MTL across heterogeneous prediction types.

Transfer learning addresses domain shifts between training and reference observations by pre-training models in one environment and adapting them later to another one~\cite{electronics10121491}. Human-subject studies show that exposure to diverse and challenging environments enhances learning and generalization~\cite{green2012learning}, a principle that motivates designing machine learning systems with varied training tasks to improve transfer and adaptation to new environments. \emph{GMADM} applies multi-task pre-training in multi-agent reinforcement learning to learn a shared decision module across different tasks and environments, enabling efficient transfer to new scenarios~\cite{WANG2025129524}. In contrast, we study transfer across within-game environments, specifically across game maps with shared rules and prediction targets, in a supervised multi-task setting using existing gameplay datasets.

\section{Data}
    \label{sec:data}

We evaluate MTL on a large-scale \emph{World of Tanks} dataset and define 10 prediction tasks spanning two common output modalities: binary classification and regression. For classification tasks, models are trained using binary cross-entropy, while regression tasks use mean squared error (MSE). The selected tasks are plausible prediction targets available in structured telemetry from team-based multiplayer games and are relevant to a range of downstream analytics applications. While some targets are related, they are not redundant: they differ in temporal scope and intended use. In particular, \textsc{Future HP} (health points) models short-term state evolution within the prediction horizon, whereas \textsc{Final HP} targets the eventual end-of-match state. Likewise, \textsc{Damage Delta} is defined separately from the health-state targets and captures combat contribution over the prediction horizon rather than simply duplicating received-damage information. Although loss functions are not necessarily optimal for every individual task, they provide a simple and consistent setting for studying task interaction, weighting, and relative performance in the multi-task regime. In summary, we use the following ten prediction tasks and evaluation metrics:

\begin{itemize}[leftmargin=*]
    \item \textsc{Win:} predicts whether the target unit's team wins the match; evaluated using AUC.
    \item \textsc{Alive:} predicts whether the target unit survives until the end of the match; evaluated using AUC.
    \item \textsc{Hurt:} predicts whether the target unit takes damage within the prediction horizon; evaluated using AUC.
    \item \textsc{Deal Damage:} predicts whether the target unit deals damage within the prediction horizon; evaluated using AUC.
    \item \textsc{Move:} predicts whether the target unit moves within the prediction horizon; evaluated using AUC.
    \item \textsc{Final HP:} predicts the target unit's final health at the end of the match; evaluated using RMSE.
    \item \textsc{Move Dist.:} predicts the target unit's movement distance within the prediction horizon; evaluated using RMSE.
    \item \textsc{Position:} predicts the target unit's 2D position coordinates at the prediction horizon; evaluated using RMSE.
    \item \textsc{Damage Delta:} predicts damage dealt by the target unit within the prediction horizon; evaluated using RMSE.
    \item \textsc{Future HP:} predicts the target unit's health at the prediction horizon; evaluated using RMSE.
\end{itemize}

\begin{figure}[t]
    \centering
    \ifarxiv
        \includegraphics[
            width=\linewidth,
            trim={0 2cm 0 0},
            clip
        ]{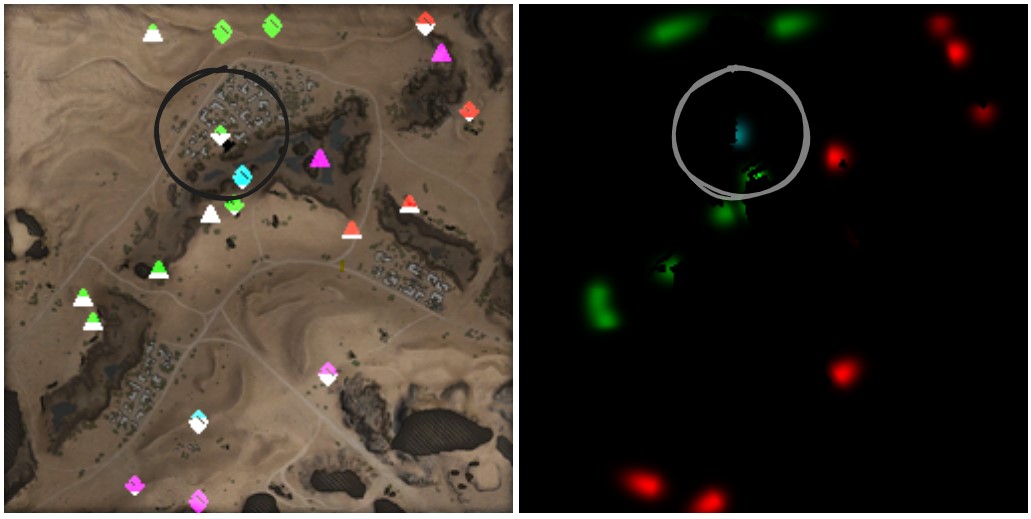}
    \else
        \includegraphics[
            width=\linewidth
        ]{figures/wot_image_input.jpg}
    \fi
    \caption{Left: Example \emph{World of Tanks} input image on the map
    \emph{Sand River}. Right: RGB mini-map background with additional
    float feature maps encoding vehicle states as Gaussian ellipsoids.}
    \label{fig:wot_image_input}
\end{figure}

\noindent The dataset comprises 2.19 million \emph{World of Tanks} battles, each involving 30 vehicles over approximately 5 minutes. Battles span 29 diverse maps of size 1{,}000 $\times$ 1{,}000\,m with an approximately equal distribution across the maps. The data was collected over 810 days and are biased toward higher-tier, more experienced players. Each game state is represented as a multimodal record combining image-based map context, global match information, and per-vehicle features and history, from which both binary classification and regression targets are derived. We use the same input representation and sampling procedure as the reference endpoint prediction model~\cite{peche2025multimodalarchitectureendpointposition}. Each sample consists of:
\begin{itemize}[leftmargin=*]
    \item \textbf{Image input:} an RGB mini-map background augmented with feature maps that encode the target vehicle, allied vehicles, and enemy vehicles as masked Gaussian ellipsoids aligned with their velocity. Instead of a sparse one-hot target pixel, the vehicles are represented as smooth, normalized heatmaps $K(x,y)=G_x(x)G_y(y)$, where $G_x$ and $G_y$ are Gaussians and the vertical component is multiplied by the shaping mask
    \begin{equation}
        M(y) =
        \begin{cases}
        \left(\dfrac{y + y_{\mathrm{mod}}}{y_{\mathrm{mod}}}\right)^{\!3}, & -y_{\mathrm{mod}} \le y < 0 \\
        1, & \text{otherwise.}
        \end{cases}
    \end{equation}
    The ellipsoid is stretched as a function of the normalized current speed and rotated into the direction of travel. Finally, obstructed regions are masked out and $K$ is normalized to sum to 1, resulting in a probability distribution over the map. Figure~\ref{fig:wot_image_input} shows an example input with the Gaussian blobs.
    \item \textbf{Global data:} a fixed-size vector containing categorical and numerical match context (map and time) as well as team-wise aggregates over vehicle healths.
    \item \textbf{Unit data:} per-vehicle categorical and numerical features (e.g., team, vehicle type/role, position, speed, health, damage, visibility-related signals) together with a short history sequence of 15 samples, each being 15 seconds apart.
\end{itemize}

\noindent Samples are drawn from 15\,s ticks of each battle. We omit timesteps that would exceed the prediction horizon and sample the horizon uniformly from 1 to 6 steps (15--90\,s). Dead vehicles are excluded from the prediction.

\section{Methodology}
    \label{sec:methodology}

To study the effect of MTL, we train a single shared encoder jointly on all tasks, attach lightweight task-specific heads for classification and regression, and compare task-balancing strategies under identical optimization, data, and evaluation protocols.
Because these heads induce very different loss scales and gradient directions, we evaluate several MTL balancing strategies (Section~\ref{subsec:loss_weighting}) for joint training. As baselines, we also report per-task single-task training (i.e., without MTL). For transfer learning, we evaluate source-task pre-training followed by target-task fine-tuning under different target-data budgets, and we also test whether the shared representation learned by MTL transfers across maps. 

\subsection{Model}

\begin{figure}
    \centering
    \includegraphics[width=1.0\linewidth,trim={0cm 0cm 0cm 0cm},clip]{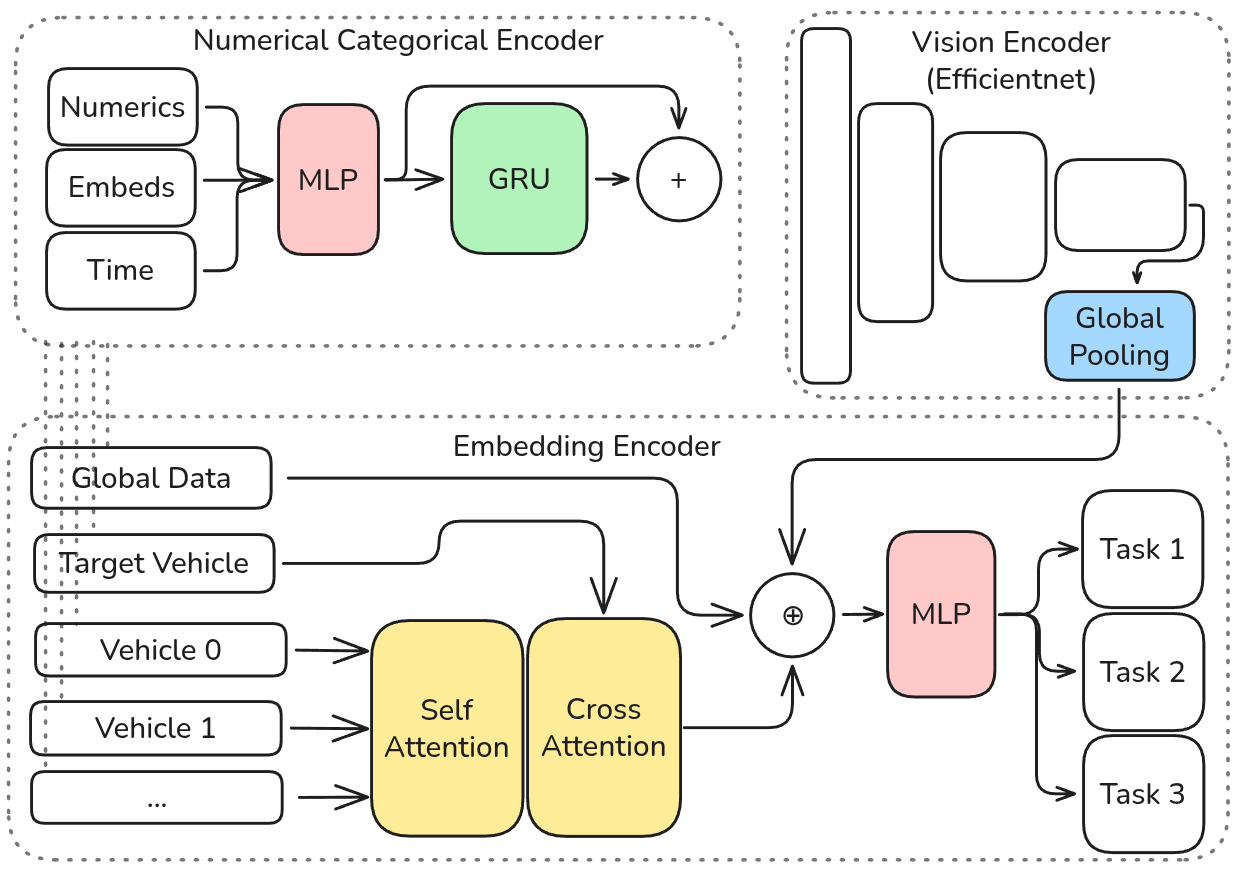}
    \caption{High-level overview of the architecture. Features of the target unit and all other units are encoded, combined through attention layers, and fused with global and visual features into the final shared embedding.}
    \label{fig:architecture}
\end{figure}

We use the multi-unit prediction architecture shown in Figure~\ref{fig:architecture}, based on prior work~\cite{peche2025multimodalarchitectureendpointposition} but without the vision decoder, as our focus is on low-dimensional classification and regression tasks rather than dense image-like outputs. A numerical-categorical encoder (NCE) maps both global data and unit specific data to embeddings. Static inputs are encoded using learned embeddings and dense layers, while dynamic sequences are encoded with a GRU~\cite{cho2014learning}. A multi-head self-attention layer models inter-unit relations. A cross-attention layer, using the target unit as the query, produces an enriched target representation, which is concatenated with the encoded and projected vision embedding and then projected to the final shared embedding. This shared embedding is used by all task-specific heads. We use attention blocks following Vaswani et al.~\cite{vaswani2023attentionneed}.

To process the rasterized map input, we use a pretrained \emph{EfficientNet-B0} model~\cite{tan2020efficientnetrethinkingmodelscaling,rw2019timm} as the vision encoder. The rasterized game-state image is passed through the network, and its 1000-dimensional output is projected to 256 dimensions and used as a fixed-size visual descriptor. We use \emph{EfficientNet-B0} as the vision encoder because it provides a practical balance between model capacity and efficiency~\cite{tan2020efficientnetrethinkingmodelscaling}. While newer encoders such as \emph{Swin Transformer}~\cite{liu2022swintransformerv2scaling} or \emph{ConvNeXt}~\cite{liu2022convnet2020s} may offer different trade-offs, our goal is to compare MTL strategies within a fixed multimodal architecture rather than to maximize backbone accuracy. We therefore decided to use a compact, established convolutional encoder.

\subsection{Loss Weighting}
    \label{subsec:loss_weighting}

Because of the very different prediction head architectures, loss functions, and resulting gradient scales, simply summing the losses is not sufficient. 
To illustrate this, Figure~\ref{fig:gradient_conflict} shows a typical situation in which gradients from different tasks point in conflicting directions. In practice, we observe three recurring issues that motivate loss weighting:

\begin{figure}[t]
    \centering
    \ifarxiv
        \def\gradientyscale{1.35}
    \else
        \def\gradientyscale{1.85}
    \fi

    \begin{tikzpicture}[
        xscale=1.85,
        yscale=\gradientyscale,
        >=stealth
    ]
        \draw[->,gray!60] (-0.2, 0) -- (3.4, 0) node[above] {$\theta_1$};
        \draw[->,gray!60] (0, -0.2) -- (0, 1.3) node[left] {$\theta_2$};

        \coordinate (O) at (2.5, 0);
        \coordinate (g1) at (0.1, 0.2);
        \coordinate (g2) at (3.0, 1.0);
        \coordinate (gs) at (1.55, 0.6); 
        
        \coordinate (g1o) at (0.5, 1.0); 
        \coordinate (g2o) at (2.586, 1.03); 
        \coordinate (go) at (1.543, 1.0); 

        \draw[->,very thick,blue] (O) -- (g1) node[above right] {$g_1$};
        \draw[->,very thick,red]  (O) -- (g2) node[below right] {$g_2$};
        \draw[->,very thick,black] (O) -- (gs) node[below left] {$\sum_i w_i g_i$};
        \draw[->,very thick,green!60!black] (O) -- (go) node[above] {$g_{\mathrm{mod}}$};
        
        \draw[->,black!60] (O) -- (g1o) node[above] {$\mathrm{proj}_{g_{\mathrm{1}}}$};
        \draw[->,black!60] (O) -- (g2o) node[above] {$\mathrm{proj}_{g_{\mathrm{2}}}$};

        \draw[gray!70] (O) ++(63.43:0.9) arc[start angle=63.43,end angle=175.24,radius=0.9];
        \node[gray!70] at (2.3, 0.7) {conflict};
    \end{tikzpicture}
\caption{Gradient interference in MTL: task gradients can differ in both direction and magnitude, leading to conflicts (blue vs.\ red) and scale mismatch. As a result, naive aggregation (black) can be suboptimal. Gradient manipulation methods such as \emph{PCGrad} mitigate these conflicts through projection, producing a better aligned update direction (green).}
    \label{fig:gradient_conflict}
\end{figure}
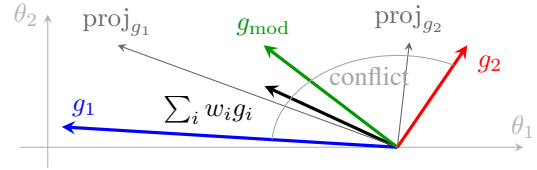

\begin{itemize}[leftmargin=*]
    \item \textbf{Scale mismatch:} different loss functions (e.g., BCE vs. MSE) produce gradients with very different magnitudes, causing one task to dominate.
    \item \textbf{Gradient conflict:} gradients may have negative cosine similarity, so optimizing one task can directly harm another.
    \item \textbf{Unbalanced progress:} even without strong conflict, some tasks converge faster and stop improving unless the optimizer reallocates capacity.
\end{itemize}

\noindent Taken together, these issues suggest that naively summing task losses is often insufficient, motivating explicit loss-weighting or gradient-balancing methods.

In general, we differentiate between loss-weighting methods, which scale each task loss by a constant or dynamic factor, and gradient-based methods, which directly modify the gradient vectors. Not all tasks are affected equally by the issues above, and we therefore evaluate a representative set of approaches as listed below. Other groups of balancing methods include task-sampling strategies, which control the effective learning signal of each task by adjusting the sampling frequency. These are particularly relevant when not every training example provides labels for every task. Further multi-task approaches address interference through architectural changes such as via soft parameter sharing~\cite{ruder2017overviewmultitasklearningdeep}, cross-stitch networks~\cite{misra2016crossstitchnetworksmultitasklearning}, or mixture-of-experts models~\cite{10.1145/3219819.3220007}. These additional groups are not considered further in this paper, as our focus is on optimization-based balancing under a fixed hard-parameter-sharing architecture. Hard parameter sharing is a natural and practical choice for our setting, where a common game-state encoding is learned jointly across several related tasks. Likewise, task-sampling methods are less relevant here because each training state provides supervision for all tasks, reducing the potential benefit of sampling-based rebalancing. We evaluate the following loss-weighting and gradient-balancing methods:

\begin{description}[leftmargin=*,topsep=0.5\baselineskip,itemsep=0.5\baselineskip]
    \item[Random Loss Weighting (RLW)] RLW is a baseline that trains an MTL model with random loss weights sampled on the probability simplex at each step.
    Weights are drawn from a symmetric Dirichlet distribution
    \begin{equation}
    \label{eq:rlw}
    \begin{aligned}
    w^{(t)} &\sim \mathrm{Dirichlet}(\alpha \mathbf{1}), \quad \alpha>0, \\
    \mathcal{L}(\theta; w^{(t)}) &= \sum_{i=1}^{m} w^{(t)}_i \, L_i(\theta).
    \end{aligned}
    \end{equation}
    which yields $\mathbb{E}[w_i^{(t)}]=\tfrac{1}{m}$ but injects controlled stochasticity into the task trade-off. Lin et al.~\cite{lin2022reasonableeffectivenessrandomweighting} demonstrate that such random weighting can match many specialized MTL weighting strategies and should therefore be treated as a key baseline.

    \item[Fast Adaptive Multitask Optimization (FAMO)] Let $L_i(\theta)$ denote the loss of task $i \in \{1,\dots,m\}$ and $g_i^{(t)} \!=\! \nabla_{\theta} L_i(\theta^{(t)})$ its gradient at step $t$. FAMO adaptively selects weights $w^{(t)} \in \Delta^{m-1}$ (the probability simplex) to encourage a \emph{balanced decrease} of all task losses while avoiding the $\mathcal{O}(m)$ time and space overhead of approaches that explicitly compute and store all per-task gradients at each step. The resulting update can be expressed as a weighted gradient descent step
    \begin{equation}
    \label{eq:famo_dir}
    g^{(t)} \;=\; \sum_{i=1}^{m} w^{(t)}_i \, g_i^{(t)},
    \qquad
    \theta^{(t{+}1)} \leftarrow \theta^{(t)} - \eta \, g^{(t)},
    \end{equation}
    where FAMO computes $w^{(t)}$ with $\mathcal{O}(1)$ overhead relative to a standard optimizer such as Adam or AdamW, reallocating weight toward tasks that are progressing more slowly to mitigate unbalanced optimization~\cite{liu2023famo}.

    \item[PCGrad] In contrast to loss-reweighting methods, which only scale $L_i$, \emph{PCGrad} is a \emph{gradient manipulation} method that directly modifies per-task gradients to reduce \emph{gradient conflict}. It performs \emph{gradient surgery} by projecting a task gradient away from other task gradients when they conflict, i.e., when their dot product is negative, and then aggregates the modified gradients for the parameter update~\cite{yu2020gradient}. We write the update in the generic form
    \begin{equation}
    \begin{split}
    \tilde{g}_i^{(t)} &= \mathrm{PCGrad}\big(g_i^{(t)}, \{g_j^{(t)}\}_{j\neq i}\big),\\
    g^{(t)} &= \frac{1}{m}\sum_{i=1}^{m} \tilde{g}_i^{(t)},\\
    \theta^{(t+1)} &\leftarrow \theta^{(t)} - \eta \, g^{(t)}.
    \end{split}
    \end{equation}
    \noindent While there are improved gradient-based alternatives such as \emph{CAGrad}~\cite{liu2021cagrad}, we use \emph{PCGrad} here for simplicity and computational efficiency.
\end{description}

\subsection{Training Details}

All experiments use a batch size of 128 and an embedding size of 128, resulting in 6.3M trainable parameters, of which 5.3M belong to the vision encoder. We run three seeded trials, using the same dataset split to reduce optimization noise. The model is trained with the \emph{AdamW} optimizer (learning rate $10^{-4}$, weight decay $10^{-3}$) and a cosine annealing schedule with restarts (initial period 100 steps, doubled after each restart, decay factor $\gamma=0.85$, minimum learning rate $10^{-6}$, and 10 warmup steps). Early stopping uses a patience of 10 epochs and monitors the unweighted mean of the validation losses across all tasks. We deliberately avoid weighted validation objectives, as these would couple model selection to the internal behavior of the balancing method itself, and we also avoid per-task checkpoint selection, since the focus of this work is a single shared model rather than task-wise model selection. The chosen stopping rule is simple and imperfect, as it does not explicitly address differences in task scale or convergence speed, but it yields a uniform and method-independent comparison protocol. In the full-data setting, validation curves were generally stable and overfitting was limited. For the fine-tuning experiments, the goal is to evaluate whether source-task or cross-map pre-training provides a useful initialization under the same model-selection protocol, rather than to maximize each task independently. FAMO uses AdamW for its inner optimizer with a learning rate of $10^{-3}$ and $\gamma=10^{-2}$. We hold out 1,000 full battles for validation and 10,000 for testing. The remaining 2.18 million battles are used for training, unless otherwise defined. Samples from the same battle never appear in different splits.

\section{Evaluation}
    \label{sec:evaluation}

We performed ablation studies under different training setups while keeping the model architecture and training parameters unchanged unless noted otherwise. For binary classification tasks, we report the area under the ROC curve (AUC), and for regression tasks, root mean squared error (RMSE). We additionally report task losses to highlight differences in loss scale and scenarios where raw training behavior matters beyond final metrics.

\subsection{RQ1: Single-Task vs. Multi-Task Performance}

We first compare single-task learning (STL) and MTL with identical architecture and optimization settings. For MTL, we use equal weighting as the main baseline. For the reported metrics, $\uparrow$ indicates that higher values are better and $\downarrow$ that lower values are better. Results are summarized in Table~\ref{tab:stl_mtl}, including the final test-set loss to highlight the difference in loss magnitudes. Numerically, EW MTL improves most tasks, except for winning prediction -- an especially long-term and hard task -- and position regression. 

\begin{table}[t]
\centering
\caption{Comparison of single-task learning (STL) and equal-weight MTL (EW MTL). STL loss is included to highlight differences in magnitude across tasks.}
\begin{tabularx}{0.95\linewidth}{ll@{\hskip30pt}ccc}
\toprule
\textbf{Group} & \textbf{Task} & \textbf{STL} & \textbf{EW MTL} & \textbf{STL Loss} \\
\midrule
\multirow{5}{*}{AUC $\uparrow$}
    & Win           & 0.8353 & 0.8210 & 0.4909 \\
    & Alive         & 0.7379 & 0.7533 & 0.5903 \\
    & Hurt          & 0.7469 & 0.7950 & 0.5344 \\
    & Deal damage   & 0.7082 & 0.7529 & 0.6066 \\
    & Move          & 0.8573 & 0.8930 & 0.2434 \\
\midrule
\multirow{5}{*}{RMSE $\downarrow$}
    & Final HP      & 0.3139 & 0.2885 & 0.0990 \\
    & Move dist.    & 0.1185 & 0.1159 & 0.0142 \\
    & Position      & 0.1159 & 0.1242 & 0.0135 \\
    & Damage delta  & 0.3320 & 0.3141 & 0.1122 \\
    & Future HP     & 0.2486 & 0.2378 & 0.0624 \\
\bottomrule
\end{tabularx}
\label{tab:stl_mtl}
\end{table}

\subsection{RQ2: MTL Loss Weighting Strategies}

We next compare different weighting and gradient-balancing strategies: equal weighting (EW), RLW, FAMO, and \emph{PCGrad}. The results are summarized in Table~\ref{tab:mtl_weighting}. From this point onward, we report the average within each metric group to provide a higher-level summary. This comparison tests whether more advanced MTL optimization strategies improve the trade-off across heterogeneous tasks.

RLW, recommended as a baseline MTL weighting method~\cite{lin2022reasonableeffectivenessrandomweighting}, shows mixed success and performs below EW in our setting. FAMO shows normalization effects on the weights, up-weighting the regression tasks but harming the classification tasks. \emph{PCGrad} provides the best aggregate numbers among the evaluated approaches, but its margin over EW is small while its training time is substantially higher, as shown in Table~\ref{tab:performance}. 

\begin{table}[t]
\centering
\caption{Comparison of different MTL weighting strategies with single-task learning as a reference.}
\begin{tabularx}{1.0\linewidth}{Xccccc}
\toprule
\textbf{Metric} & \textbf{STL} & \textbf{EW} & \textbf{RLW} & \textbf{FAMO} & \textbf{PCGrad} \\
\midrule
AUC $\uparrow$ & 0.7771 & 0.8030 & 0.7717 & 0.7078 & 0.8044 \\
RMSE $\downarrow$ & 0.2258 & 0.2161 & 0.2211 & 0.2204 & 0.2159 \\
Loss $\downarrow$ & 0.2767 & 0.2632 & 0.2770 & 0.3002 & 0.2626 \\
\bottomrule
\end{tabularx}
\label{tab:mtl_weighting}
\end{table}

\subsection{RQ3: Pre-training and Low Data}

So far, we have relied on a large dataset. In production, however, data is often scarce. In such settings, MTL may improve stability and performance through knowledge transfer. To test this, we split the tasks into two groups. The six source tasks used for \emph{PCGrad}-based multi-task pre-training are: \textsc{Hurt}, \textsc{Deal Damage}, \textsc{Move}, \textsc{Move Dist.}, \textsc{Position}, and \textsc{Damage Delta}. The four target tasks used for downstream adaptation are: \textsc{Win}, \textsc{Alive}, \textsc{Final HP}, and \textsc{Future HP}. We then compare models trained from scratch on varying amounts of data (100, 1,000, 10,000, and the full 2.18 million) for the target tasks with models pre-trained on the full dataset using only the source-task labels and subsequently fine-tuned on the target tasks. Newly introduced task-specific fully connected layers are randomly initialized using \emph{PyTorch}'s default \texttt{nn.Linear} initialization, while pre-trained backbone components retain their transferred weights. To mitigate overfitting in this low-data regime, we reduce both the learning rate and the minimum learning rate by a factor of 10 for all trainings not performed on the full data. We keep the architecture fixed across all data set sizes to preserve comparability, even though this likely leaves the smallest-data experiments in an overparameterized setting. Table~\ref{tab:pretrain_finetune} shows positive knowledge transfer across all reported aggregate metrics, with the gains tending to decrease as more target-task data becomes available. This setup should be interpreted as a practical transfer-learning scenario with abundant source-task data and limited target-task labels, rather than as a controlled comparison with matched total data volume.

\begin{table}[t]
\centering
\caption{Performance of training from scratch and fine-tuning across dataset sizes. Data quantity is given in full battles.}
\begin{tabularx}{0.9\linewidth}{ll@{\hskip30pt}ccc}
\toprule
\textbf{Data} &\textbf{Metric} & \textbf{Scratch} & \textbf{Fine-tune} & \textbf{Rel. Change} \\
\midrule
\multirow{3}{*}{100}
  & AUC $\uparrow$ & 0.5525 & 0.5659 & +2.42\% \\
  & RMSE $\downarrow$ & 0.3477 & 0.3178 & -8.59\% \\
  & Loss $\downarrow$ & 0.4048 & 0.3906 & -3.50\% \\
\midrule
\multirow{3}{*}{1000}
  & AUC $\uparrow$ & 0.5814 & 0.5983 & +2.91\% \\
  & RMSE $\downarrow$ & 0.3272 & 0.2957 & -9.62\% \\
  & Loss $\downarrow$ & 0.3915 & 0.3795 & -3.07\% \\
\midrule
\multirow{3}{*}{10000}
  & AUC $\uparrow$ & 0.5929 & 0.6057 & +2.16\% \\
  & RMSE $\downarrow$ & 0.3090 & 0.2928 & -5.23\% \\
  & Loss $\downarrow$ & 0.3837 & 0.3765 & -1.88\% \\
\midrule
\multirow{3}{*}{Full}
  & AUC $\uparrow$ & 0.7448 & 0.7514 & +0.88\% \\
  & RMSE $\downarrow$ & 0.2720 & 0.2678 & -1.53\% \\
  & Loss $\downarrow$ & 0.3240 & 0.3230 & -0.33\% \\
\bottomrule
\end{tabularx}
\label{tab:pretrain_finetune}
\end{table}

\subsection{RQ4: Multi-Map Training}

Another potential source of generalization in video games is variation across different environments such as game maps. In our setting, maps preserve the same prediction targets, unit types, and mechanics but change the input distribution through terrain layout, visibility, and movement affordances. Cross-map transfer is therefore a within-game environment shift, not evidence of transfer across different games or rule systems. It is nevertheless relevant in commercial games where training data on a newly released or unreleased map is often limited. In such cases, transfer learning from large-scale datasets collected on existing maps is highly relevant. We therefore compare training on a single target map with prior pre-training on all other maps (which cover a broad range of environment layouts, including open terrain, covered routes, and built-up areas) and evaluate whether such cross-map pre-training improves generalization on the target map. As target environment, we choose \emph{Redshire}, a map with a mixed terrain profile that combines open terrain, cover, and scattered  built-up areas, making it a representative heterogeneous target map. As in the previous section, we compare several target-data regimes, ranging from small subsets to the full target-map dataset. 

The corresponding results are given in Table~\ref{tab:multi_env}. Across all target-data regimes, pre-training on other maps improves downstream performance on \emph{Redshire}, with the gains decreasing as more target-map data becomes available. This behavior is consistent with a transfer-learning effect in which the shared representation learned from multiple tasks and maps provides a stronger initialization in low-data settings. Because the cross-map pre-training setup also uses substantially more source data, these results should be interpreted as evidence for the practical value of transfer from existing maps, rather than as a controlled measurement of the effect of map diversity alone.

\begin{table}[t]
\centering
\caption{Performance on the map \emph{Redshire} across dataset sizes when trained from scratch and when fine-tuned from a pretrained base model. Data quantity is given in full battles.}
\begin{tabularx}{0.9\linewidth}{ll@{\hskip30pt} ccc}
\toprule
\textbf{Data} & \textbf{Metric} & \textbf{Scratch} & \textbf{Fine-tune} & \textbf{Rel. Change} \\
\midrule
\multirow{3}{*}{100}
  & AUC $\uparrow$ & 0.6288 & 0.7591 & +20.72\% \\
  & RMSE $\downarrow$ & 0.2627 & 0.2227 & -15.24\% \\
  & Loss $\downarrow$ & 0.3289 & 0.2797 & -14.96\% \\
\midrule
\multirow{3}{*}{1000}
  & AUC $\uparrow$ & 0.6792 & 0.7812 & +15.03\% \\
  & RMSE $\downarrow$ & 0.2408 & 0.2176 & -9.62\% \\
  & Loss $\downarrow$ & 0.3102 & 0.2693 & -13.17\% \\
\midrule
\multirow{3}{*}{10000}
  & AUC $\uparrow$ & 0.6908 & 0.7843 & +13.54\% \\
  & RMSE $\downarrow$ & 0.2376 & 0.2166 & -8.84\% \\
  & Loss $\downarrow$ & 0.3060 & 0.2678 & -12.48\% \\
\midrule
\multirow{3}{*}{Full}
  & AUC $\uparrow$ & 0.8044 & 0.8144 & +1.24\% \\
  & RMSE $\downarrow$ & 0.2109 & 0.2087 & -1.01\% \\
  & Loss $\downarrow$ & 0.2591 & 0.2544 & -1.84\% \\
\bottomrule
\end{tabularx}
\label{tab:multi_env}
\end{table}

\subsection{Performance Considerations}

In production settings, memory footprint as well as training and inference cost are often important constraints. Jointly training several tasks with a shared model can reduce these costs compared to maintaining separate STL models. Table~\ref{tab:performance} summarizes the differences in parameter count, floating-point operations, and training time between 10 separate STL models and a single MTL model, trained with EW or with \emph{PCGrad}. Since \emph{PCGrad} depends on separate gradient operations per task, training time scales with the number of tasks. These numbers compare 10 same-capacity STL models with one shared MTL model; they do not prove that each STL model is parameter-optimal. EW MTL provides the clearest efficiency benefit, while \emph{PCGrad} trades much of the training-time saving for a small numerical gain in aggregate metrics. 

\begin{table}[t]
\setlength{\tabcolsep}{4pt}
\centering
\caption{Efficiency comparison in terms of parameter count, floating-point operations, and training time.}
\label{tab:performance}
\begin{tabularx}{1.0\linewidth}{Xccc}
    \toprule
    \textbf{Setup} & \textbf{Params} & \textbf{GFLOPs/} & \textbf{Training Time} \\
                   &                 & \textbf{forward pass}     & \\
    \midrule
    10$\times$ STL models & 63.0M & 10.29  & 2:27h \\
    1$\times$ MTL model (EW) & 6.3M & 1.03 & 1:36h \\
    1$\times$ MTL model (PCGrad) & 6.3M & 1.03 & 4:35h \\
    \bottomrule
\end{tabularx}
\end{table}

\section{Limitations and Future Work}

This paper has several limitations and directions for future work. First, we report mean results without confidence intervals or formal significance tests, so small differences between methods should be interpreted with caution. Another limitation is the use of a simple early-stopping mechanism, which becomes less suitable for smaller datasets. Monitoring a single joint value neither reflects the progress of individual tasks nor can it fully prevent overfitting. Even a better aggregate metric than the average loss would only balance between those two issues. More robust alternatives could freeze parts of the model, such as the encoder, and then apply task-specific early stopping or gradually freeze individual heads.

In this paper, we explored source-task and cross-map pre-training as means to transfer knowledge from larger labeled datasets, but one side effect is a relatively large model, which may exceed the inference-time budget for downstream tasks. Distillation is a popular solution to extract patterns and knowledge using a large teacher model and then use it to train a smaller model.

For the sake of simplicity, \emph{PCGrad} was used as the weighting algorithm for the pretraining/fine-tuning experiments throughout this paper. There are many variations and similar, yet distinct, approaches for balancing task weighting methods (such as \emph{SLGrad} or \emph{CAGrad}). More state-of-the-art approaches should be tested with this paper's combination of model architecture and dataset properties.

Since the effect of MTL strongly correlates with the dataset and its environmental and statistical properties, additional datasets such as \emph{ESTA}~\cite{xenopoulos2022estaesportstrajectoryaction} should be considered in future evaluations, together with independently optimized STL baselines and controlled cross-map transfer studies with different environments and matched data budgets. 

\section{Conclusions}

This paper investigated MTL for prediction tasks derived from structured video game state, using a large-scale \emph{World of Tanks} dataset with heterogeneous targets spanning binary classification and regression. We adapted a shared multimodal architecture that combines image input, global match information, and per-unit features, and evaluated whether joint training can improve generalization compared to single-task learning while reducing the need for separate specialized models.

Our results show that MTL is effective in this setting (RQ1), but its success depends on the optimization strategy (RQ2). \emph{RLW} does not reliably balance the heterogeneous objectives, whereas gradient-aware optimization, in particular \emph{PCGrad}, produced the best overall trade-off across tasks in our experiments (RQ2). In addition, pre-training and fine-tuning improved performance in low-data target-task settings, indicating that source-task pre-training provides a useful initialization for downstream adaptation (RQ3). Positive transfer in low-data settings and across maps suggests that multi-task supervision can improve generalization beyond the immediate training scenario, including under within-game map shifts (RQ4).

Taken together, these findings support MTL as a practical approach for supervised game-state modeling, especially when multiple related labels are available for the same observation and efficiency is important in production. At the same time, the experiments highlight that heterogeneous task sets introduce substantial optimization challenges, making balancing and checkpoint selection critical design choices. We therefore view MTL not as a drop-in replacement for single-task training, but as a promising training paradigm whose benefits depend on careful optimization, evaluation, and a clear definition of the intended transfer setting.

\iftrue
\section*{Acknowledgments}

We thank \emph{Wargaming} for providing a high-quality dataset for experimentation and evaluation.
\fi

\bibliographystyle{IEEEtran}
\bibliography{references}

\end{document}